%% file: main.tex
\documentclass[runningheads]{llncs}
\usepackage[T1]{fontenc}
\usepackage{graphicx}
\usepackage{amsfonts}
\usepackage[colorlinks=true, allcolors=blue]{hyperref}
\usepackage{cleveref}
\usepackage[inline]{enumitem}
\usepackage{siunitx}
\usepackage{booktabs,tabularray}
\usepackage{multirow}   
\usepackage{multicol} 
\usepackage{subcaption}
\usepackage{float}
\usepackage{nicefrac}
\usepackage[ruled,vlined,noend]{algorithm2e}
\usepackage{xcolor}
\usepackage{url}
\usepackage{tikz, pgfplots}
\usetikzlibrary {arrows.meta} 

\definecolor{tealtrain}{RGB}{60,71,76}
\definecolor{greenvalid}{RGB}{0,255,0} 
\definecolor{darkyellow}{RGB}{240,212,59}
\definecolor{purple}{RGB}{228,32,241}

\begin{document}
\title{Large-image Object Detection for Fine-grained Recognition of Punches Patterns in Medieval Panel Painting}
\titlerunning{Object Detection for Fine-grained Recognition of Punches Patterns}
%

\author{Josh Bruegger\inst{1}\orcidID{0009-0009-6260-3038} \and
Diana Ioana Cătană\inst{1}\orcidID{0000-0000-0000-0000}
Vanja Macovaz\inst{2}\orcidID{0000-0002-7775-8165} \and
Matias Valdenegro-Toro\inst{1}\orcidID{0000-0001-5793-9498} \and
Matthia Sabatelli\inst{1}\orcidID{0009-0007-7540-8616} \and 
Marco Zullich\inst{1}\orcidID{0000-0002-9920-9095}}
%

\authorrunning{J.\ Bruegger et al.}
%
\institute{
Faculty of Science and Engineering, University of Groningen, Nijenborgh 9, 9747 AG Groningen, the Netherlands\\
\email{marco.zullich@gmail.com}
\and
Independent researcher and art photographer}


%
\maketitle              
\begin{abstract} 
The attribution of the author of an art piece is typically a laborious manual process, usually relying on subjective evaluations of expert figures.
However, there are some situations in which quantitative features of the artwork can support these evaluations.
The extraction of these features can sometimes be automated, for instance, with the use of Machine Learning (ML) techniques.
An example of these features is represented by repeated, mechanically impressed patterns, called \emph{punches}, present chiefly in 13th and 14th-century panel paintings from Tuscany.
Previous research in art history showcased a strong connection between the shapes of punches and specific artists or workshops, suggesting the possibility of using these quantitative cues to support the attribution.
In the present work, we first collect a dataset of large-scale images of these panel paintings.
Then, using YOLOv10, a recent and popular object detection model, we train a ML pipeline to perform object detection on the punches contained in the images.
Due to the large size of the images, the detection procedure is split across multiple frames by adopting a sliding-window approach with overlaps, after which the predictions are combined for the whole image using a custom non-maximal suppression routine.
Our results indicate how art historians working in the field can reliably use our method for the identification and extraction of punches.

\end{abstract}
\section{Introduction}\label{sec:intro}

The process of attributing the author or authors of a work of art is topical within art history.
It is usually conducted by means of meticulous qualitative investigations aimed at assessing aspects such as style, perceptive visual features, and other information, such as geographical location and historical context.
However, there are examples of quantitative methods used in the process.
For instance, the study of material composition via non-destructive testing techniques like infrared thermography or x-rays can reveal specific structural features of the wooden panel of panel paintings \cite{Yao2018ActiveTT}.
These features can subsequently be used as cues to aid art historians in the attributions.

Starting from the 13\textsuperscript{th} century, panel painters in Florence and its surroundings started decorating the gold foil in their art pieces by means of mechanical tools, called \emph{punches}, that, when impressed on the gold foil, would produce a small pattern with a specific shape.
Starting from the 1960s, art historians Mojmir Frinta \cite{frinta1972punched} and Erling S.~Skaug \cite{skaug1983punch} started investigating the potential of studying these patterns in a quantitative way to draw connections between art pieces.
It is indeed very probable that a punch pattern is \emph{unique}, given that the limited technology at the time made it highly difficult to reproduce punches with the exact same shape as another one.
Specifically, Skaug conducted a very extensive investigation, manually cataloging a very large number of punches, their exact measurements, the art pieces they are connected to, and the potential author(s) involved in their production \cite{skaug1994punch}.
This work took him more than 30 years and is still not exhaustive of all the panel paintings making use of punched decoration in that geographical location.
All this considered, this process could largely benefit from the application of automatic tools that allow for the extraction and the subsequent automatic classification of the punch category, which would relieve art historians from the lengthy procedure of manual measurements and cataloging.

Motivated by the advances during the last decade in automatic image classification and object detection (OD), in the present work, we 
\begin{enumerate*}[label=(\alph*)]
    \item introduce a dataset composed of 8 ultra-high resolution images of panel paintings from Museo Nazionale di Pisa (Italy) and
    \item train a Deep Learning (DL) pipeline for performing OD on this dataset.
\end{enumerate*}
All of these paintings include examples of punched decoration from a limited set of authors, with many of the punch categories occurring in more than one work of art.
This application is challenging, given the large spatial size of the images, coupled with the relatively small dimension of the punchmarks.
This would require an unfeasible amount of computational power to be able to run ML models on the full height and width of the images.
Initially, we train YOLOv10 OD models \cite{wang2024yolov10} to jointly predict the location and classification of the punches on random crops of these panel paintings.
During inference, we tackle the computational issues by adopting a sliding window approach inspired by similar techniques in the field of computer vision (e.g., \cite{maree2004generic,nanni2014ensemble}).
We divide the images into several \emph{frames} with partial overlap.
We run each of these frames on the trained YOLOv10 model, getting a list of candidate predictions.
Finally, we combine these predictions using a custom non-maximal suppression (NMS) strategy on the overlaps between frames, getting a definitive set of predictions for the whole image.
Our approach records a Precision of 94\% and an F1-Score of 90\% on held-out data, showing how it can serve as a reliable helper tool for aiding the work of art historians in support of the attribution process.

Our data and implementation are available at the following URL: \url{https://github.com/marcozullich/punches-object-detection}.


\subsection{Related Work}\label{sec:related}

\subsubsection{Object Detection}\label{sec:related_od}
OD is one of the fundamental tasks of computer vision.
It consists of recognizing and localizing instances of known objects in images.
The literature distinguishes between two main categories of OD methods:
\begin{enumerate*}[label=(\alph*)]
    \item two-shot methods, which first identify image patches containing known objects, then perform classification on the patches and
    \item one-shot methods, which jointly perform localization and recognition at the same time.
\end{enumerate*}
Famous two-shot methods include the Region-based CNN methods \cite{girshick2014rich,ren2016faster}, while notable one-shot methods include the CNN-based YOLO \cite{redmon2016you} and its subsequent variants RetinaNet \cite{ross2017focal} and the attention-based DETR \cite{carion2020end}.
A \emph{classical} paradigm was seeing one-shot methods as faster but more inaccurate and two-shot methods as slower but more accurate \cite{ansari2021survey,carranza2020performance}.
However, recent advances caused the accuracy gap between the two to close, while one-shot methods still prove to be more efficient \cite{ren2023strong}.
The adoption of YOLOv10 \cite{wang2024yolov10}, a one-stage object detector based on YOLO, is advantageous considering the good trade-off between accuracy and time efficiency in a situation like ours, whereas the OD model has to be run on multiple frames of very large images.

\subsubsection{Machine Learning for analyzing artworks}\label{sec:related_ml_art}

ML has been applied to analyze artworks since the late 1990s, with works from Hachimura \cite{hachimura1996retrieval} and Corridoni et al.\ \cite{corridoni1996visual}.
They used classical computer vision techniques to extract features useful for information retrieval systems. 
For what concerns ML-assisted artwork attribution,
Kröner and Lattner \cite{kronerAuthenticationFreeHand1998}, and Melzer et al.\ \cite{melzerStrokeDetectionBrush1998} concentrating on the topic of authorship attribution.
These initial attempts were making use of basic feature engineering and \emph{shallow} feed-forward neural networks.
Later approaches include a mixture of unsupervised and supervised approaches---such as Hidden Markov Models, Support Vector Machines, and Clustering---for artist classification \cite{johnsonImageProcessingArtist2008}, and image descriptors to establish stylistic similarities \cite{shamirComputerAnalysisReveals2012}.
The last decade has seen an increase in the usage of DL applied to art:
David and Netanyahu \cite{davidDeepPainterPainterClassification2016} used features derived from a deep autoencoder to build an ML pipeline for author classification, while \cite{cetinic2018fine} solved the same task using feature extracted from a pre-trained Convolutional Neural Network (CNN).
Other works, such as the one by \cite{bar2015classification}, use DL for style classification.
Other approaches targeting artwork classification are reviewed by Santos et al.\ in their survey \cite{santosArtificialNeuralNetworks2021}.

More related to the present work are approaches aimed at identifying specific instances of known objects in paintings.
Despite the appeal that an end-to-end automated author classification may pose, the opacity in the decision rules operated by the model may represent a hurdle for explaining a prediction to an expert in the field, such as an art historian.
Models that instead target the presence of specific objects, such as specific people or punchmarks, can potentially be of better usage as they detect meaningful semantic features.
For instance, Seguin et al.\ \cite{seguin2016visual} used pre-trained CNNs to identify occurrences of common objects, such as people or animals, in paintings, while Gonthier et al.\ \cite{gonthier2022multiple} did so employing the OD model Faster R-CNN \cite{ren2016faster}.
Milani and Fraternali \cite{milanifedericoDatasetConvolutionalModel2021} published a dataset and a CNN-based approach for classifying depictions of saints across various paintings. Other works concentrate on recognizing specific figures, such as Leonardo Da Vinci \cite{tyler2012search} or Jesus Christ \cite{gonthier2022multiple}, or objects such as musical instruments \cite{sabatelli2018deep}.
What many of these works have in common is the fact that the subjects of the detection are people, animals, or everyday objects \cite{bengamraComprehensiveSurveyObject2024}, which may not necessarily offer strong evidence in the authentication process.
Despite not specifically performing OD, Lettner et al.\ \cite{lettner2004texture}, by performing recognition of painted strokes on drawings, recognized features that can functionally be useful for the manual process of authorship identification.
Finally, Zullich et al.\ \cite{zullich2023artificial} performed classification on a small dataset of punches images cropped from four pictures of panel paintings.
Our work is substantially different from this one since
\begin{enumerate*}[label=(\alph*)]
    \item they performed image classification, while we operate OD on full-resolution images---a task which is much more challenging---
    \item their dataset contains fewer paintings (4) while ours contains 8, and
    \item they did not extensively test their model on held-out data but rather on a subset of punches randomly obtained from the same data distribution of the training set.
\end{enumerate*}

\paragraph{Contributions}

The contributions of the present work are the following:
\begin{itemize}
    \item We train a pipeline for OD using an overlapping sliding window approach on very high-resolution images of panel paintings for punchmark recognition and localization, whereas previous works stopped at the level of image classification, and
    \item We propose a novel and effective NMS method for coalescing redundant high-confidence predictions which result after merging predictions from multiple windows.
\end{itemize}

\section{Materials and Methods}

\subsection{Dataset}\label{sec:dataset}

\begin{table}[t]
  \centering
  \caption{List of the panel paintings used to compose the dataset.
  All paintings were selected in collaboration with experts in the field, based on connections between the authors and the overlap between categories of punchmarks present in the artworks.}
  \label{tbl:paintings}
  \begin{tblr}{
      width = \linewidth,
      colspec = {Q[252]Q[340]Q[225]Q[54]},
      hline{1,10} = {-}{0.08em},
      hline{2} = {-}{0.05em},
    }
    Artist                          & Title                                     & Part                        & Year (circa)  \\
    Turino Vanni                    & Baptism of Christ                         & Whole                       & 1390   \\
    Master of Universitas Aurificum & Madonna and Child “Universitas Aurificum” & Whole                       &       \\
    Giovanni di Nicola              & Madonna and Child                         & Whole                       & 1340  \\
    Cecco di Pietro                 & Crucifixion/Eight Saints                  & Whole                       & 1386   \\
    Francesco Traini                & St. Dominic/Scenes from his life          & Top part of centre panel    & 1345  \\
    Francesco Traini                & St. Dominic/Scenes from his life          & Bottom part of centre panel & 1345  \\
    Francesco Traini                & St. Dominic/Scenes from his life          & Left side panels            & 1345  \\
    Francesco Traini                & St. Dominic/Scenes from his life          & Right side panels           & 1345  
    \end{tblr}
\end{table}

\begin{figure}[t]
    \centering
    \includegraphics[width=1.0\linewidth]{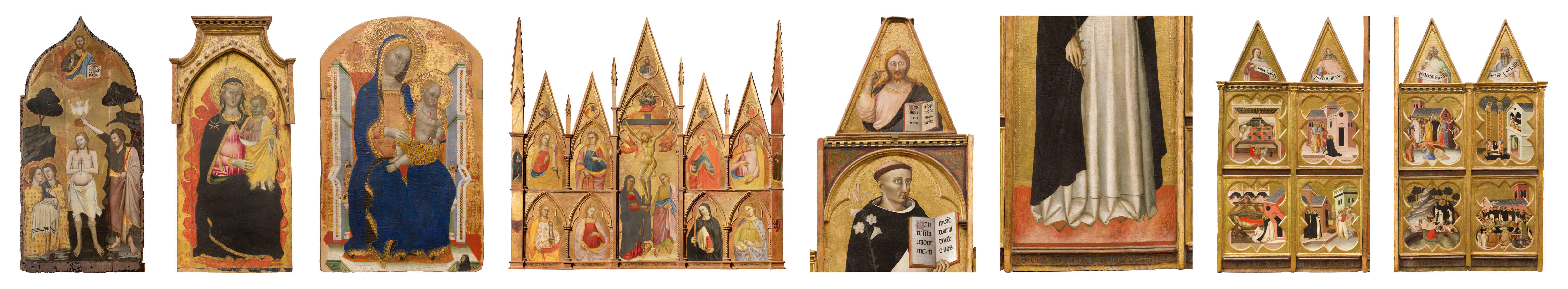}
    \caption{Composition depicting the 8 artworks composing the dataset.
    The pictures represent the paintings at a variable scale.}
    \label{fig:paintings}
\end{figure}

\begin{figure}[t]
    \centering
    \includegraphics[width=\linewidth, trim={0 10cm 0 10cm},clip]{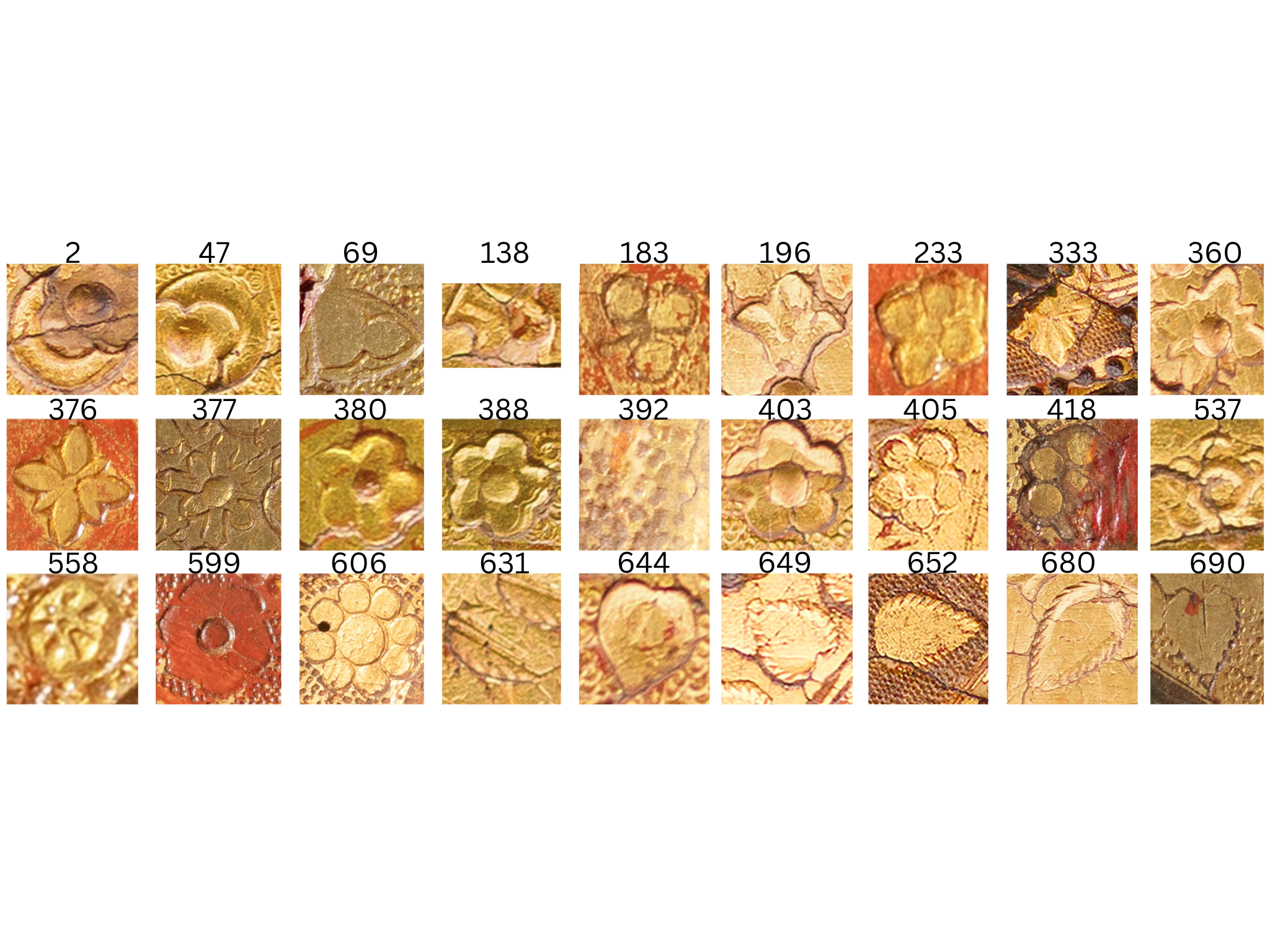}

    \caption{Samples from the punches in our dataset, one per category.}
    \label{fig:punches}
\end{figure}
    
\begin{figure}[t]
     \centering
     \begin{subfigure}[b]{0.24\linewidth}
         \centering
         \includegraphics[width=\linewidth]{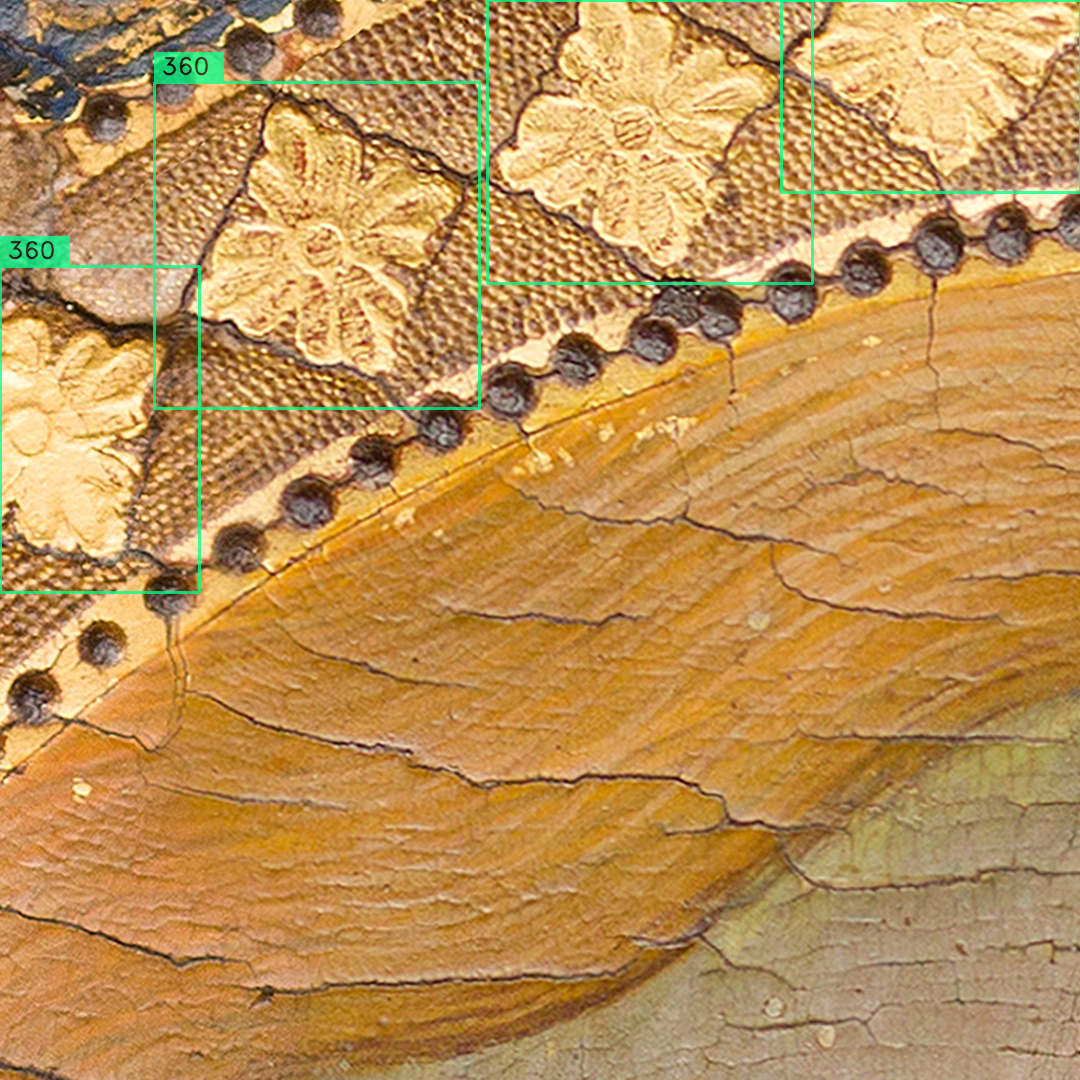}
     \end{subfigure}
     \hfill
     \begin{subfigure}[b]{0.24\linewidth}
         \centering
         \includegraphics[width=\linewidth]{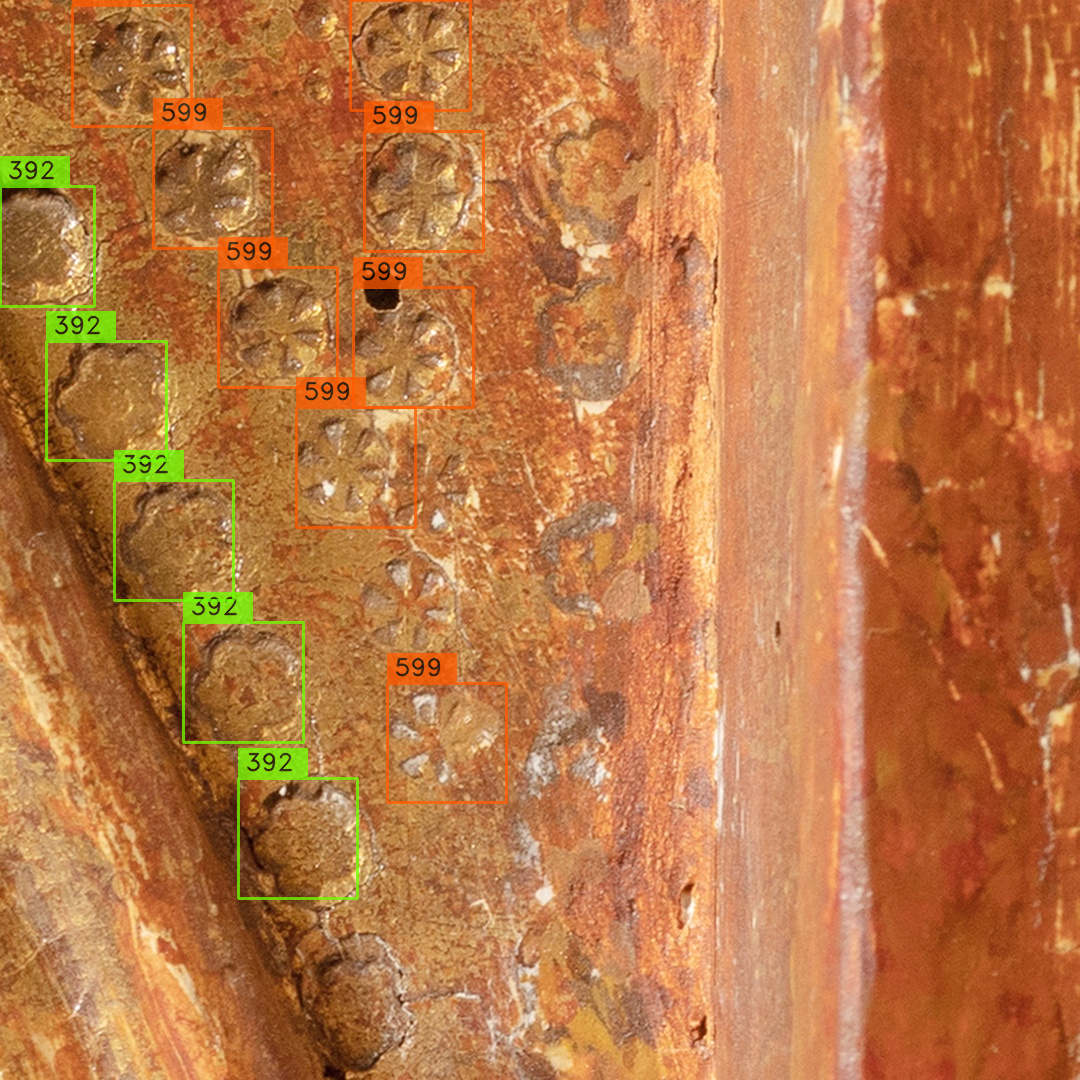}
     \end{subfigure}
     \hfill
     \begin{subfigure}[b]{0.24\linewidth}
         \centering
         \includegraphics[width=\linewidth]{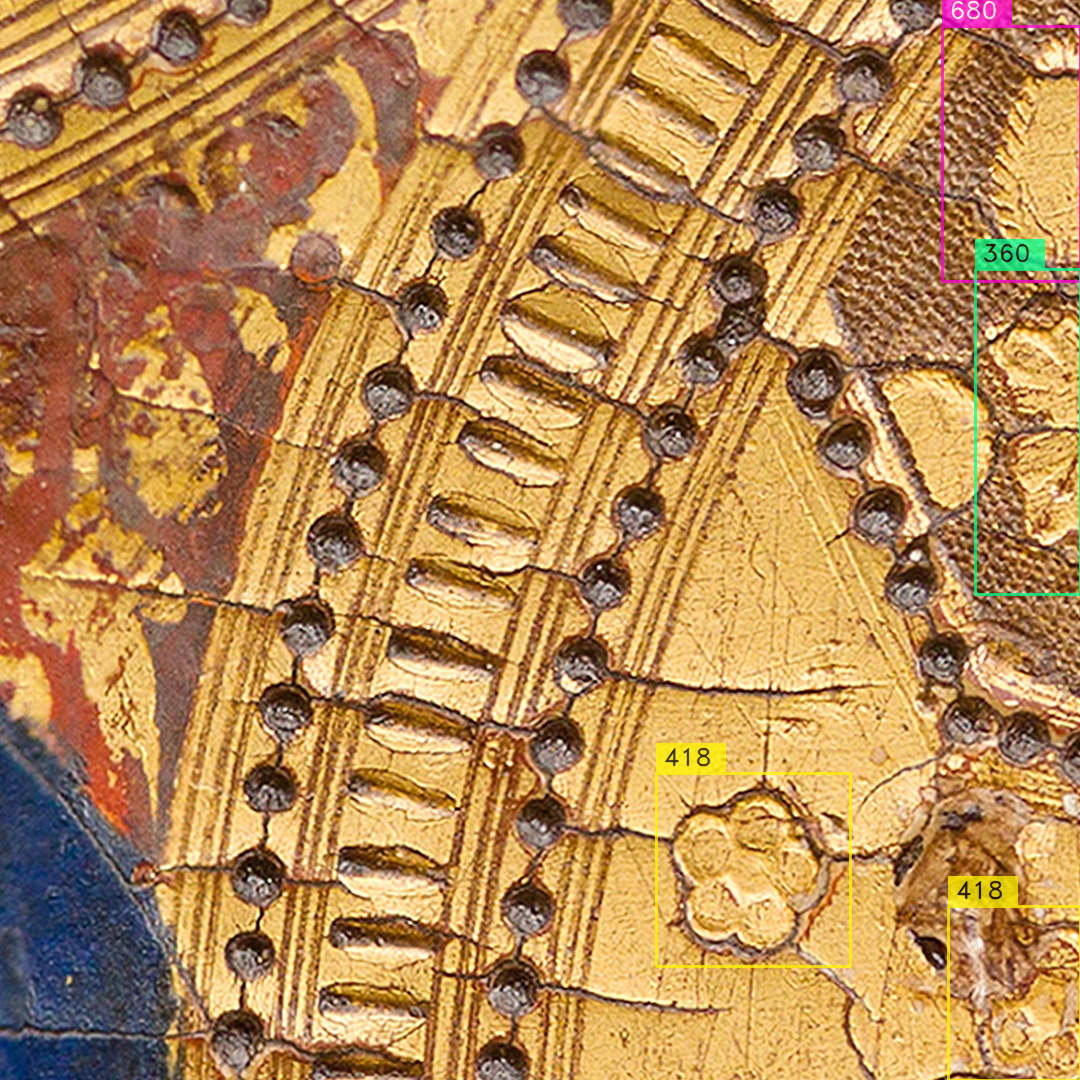}
     \end{subfigure}
     \hfill
     \begin{subfigure}[b]{0.24\linewidth} 
         \includegraphics[width=\linewidth]{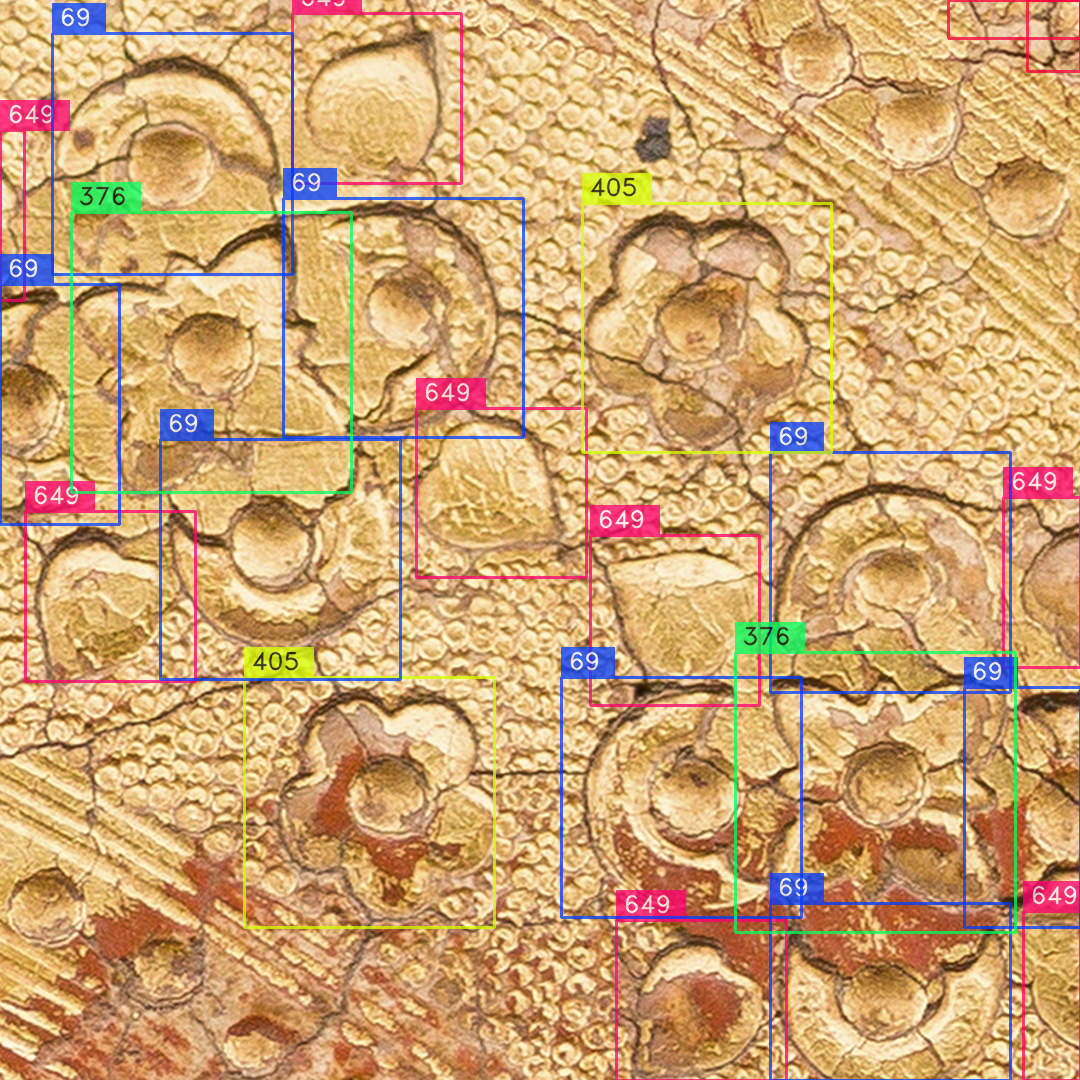}
     \end{subfigure}
        \caption{Crops of the high-resolution images of paintings showcasing some of the punchmarks after the labelling procedure.
        }
        \label{fig:crops}
\end{figure}

The dataset used in the present work is composed of 8 high-resolution pictures of panel paintings from Museo Nazionale in Pisa (Italy).
These artworks are listed in \Cref{tbl:paintings} and depicted in \Cref{fig:paintings}.
As the goal was to create a dataset with a heterogeneous set of punchmarks, but with certain instances of punches appearing in multiple art pieces, we operated the selection of paintings in collaboration with experts in the field.
We conducted the process of collecting and digitizing the paintings following the procedure indicated by Zullich et al.\ \cite{zullich2023artificial}, thus allowing us to get high-quality pictures where
\begin{enumerate*}[label=(\alph*)]
    \item the punchmarks are clearly visible in a good enough detail, and
    \item the size of the punchmarks is approximately the same for each instance, thus relieving the object detector of the task of learning a proportion invariance between instances of the same category.
\end{enumerate*}
The resulting dataset is composed of pictures of very large size (some of the images have more than \num{50000} px per side), with the smallest instances of punchmarks having just less than 100 px in resolution.
Using Adobe Photoshop, we then manually labelled the images by drawing bounding boxes around all instances of \num{3475} punchmarks distributed over 27 categories.
\Cref{fig:punches} showcases a crop for one punch mark from each of the categories; \Cref{fig:crops} instead shows some selected crops of images containing combinations of multiple punches.

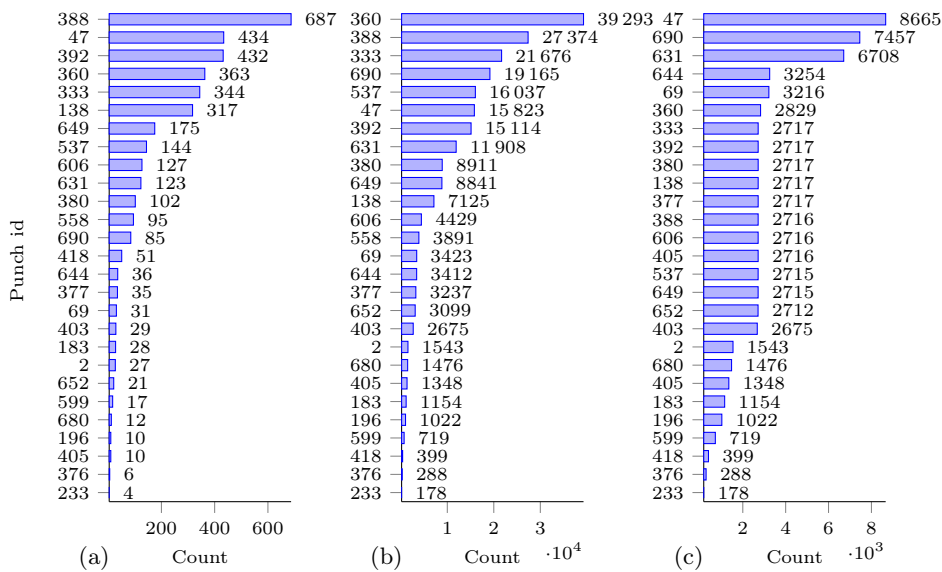
\begin{figure}
    \hspace{-0.5cm}
    \begin{subfigure}{0.25\textwidth}
        \input{plots/data_barchart}
    \end{subfigure}
    \hspace{1.25cm}
    \begin{subfigure}{0.25\textwidth}
        \input{plots/data_barchart_before_rebalancing}
    \end{subfigure}
    \hspace{0.75cm}
    \begin{subfigure}{0.25\textwidth}
        \input{plots/data_barchart_after_rebalancing}
    \end{subfigure}
    \vspace*{-1.5em}
    \caption{
    Barcharts depicting the per-category distribution of the punchmarks in our dataset.
    (a) The distribution of the original dataset before preprocessing.
    (b) The distribution after the preprocessing and before rebalancing.
    (c) The distribution after rebalancing.
    }
    \label{fig:punches_distribution}
\end{figure}

We assigned each instance to the corresponding punch, identified by the sequential number defined by Skaug in his works \cite{skaug1983punch,skaug1994punch}.
We provide a list of the punchmarks with their distribution in the dataset in \Cref{fig:punches_distribution}a.

\subsubsection{Dataset preprocessing and train-test splitting}

\begin{figure}[t]
    \centering
    \includegraphics[width=0.6\linewidth]{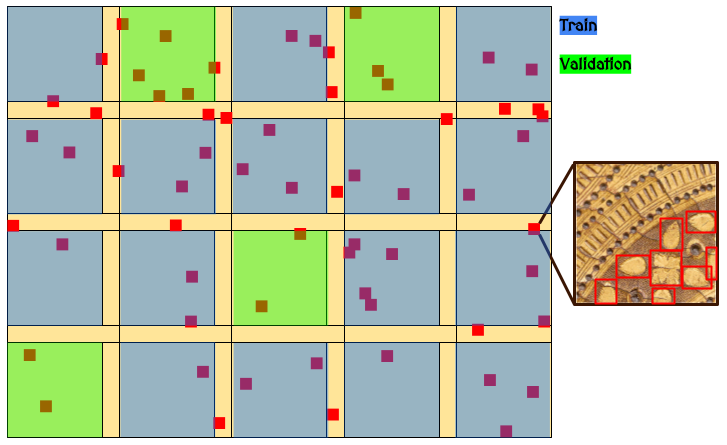}
    \caption{Illustration of the procedure we operated for splitting the dataset into training and validation splits.
    We divide the image into square grids of equal size (at least 2160 px per side, depending on the full-resolution image size).
    The frames are separated by \emph{gutters} (depicted in \textcolor{darkyellow}{\textbf{yellow}}) in order to avoid a single frame to \emph{leak} onto two different data splits.
    The cells coloured in \textcolor{blue}{\textbf{blue}} are assigned to the training set, while those depicted in \textcolor{greenvalid}{\textbf{green}} are allocated to the validation set.
    The red squares represent a possible configuration of frames obtained by the random sampling procedure.
    }
    \label{fig:data_split}
\end{figure}

Typical OD datasets contain a high number of small-resolution images with few instances of known objects each.
Our dataset, conversely, contains a small number of very high-resolution images presenting a very large number of instances of punchmarks.
In order to tackle the problem of the high resolution, we decided to crop 7 of the paintings into a total of \num{70000} frames of size \num{1088} $\times$ \num{1088} px, allowing for possible overlaps between frames.
We treated the 8\textsuperscript{th} painting as a held-out example for eventually testing the OD models.
The subdivision into frames allowed us to tackle the computational overhead represented by the high-resolution images while allowing us to obtain a much larger number of pictures.
In order to split the data between training and validation splits, we divided the images into grids, assigning given grids to the training or validation dataset.
We then randomly sampled frames within these grids, keeping a proportion of 80:20 between the two splits.
The procedure is illustrated in \Cref{fig:data_split}.

In summary, we split our dataset into a training set containing \num{56000} frames and a validation set of \num{14000} frames, each containing at least one punch mark.
The per-category distribution is presented in \Cref{fig:punches_distribution}b.
Finally, our test set is composed of one painting of dimension \num{36451}$\times$\num{27274} and containing 760 punch mark instances across four different categories, which will be used to test our trained object detector in a sliding-window fashion.

\subsubsection{Dataset rebalancing}
Considering \Cref{fig:punches_distribution}b, we can notice that there are some categories with a strong underrepresentation---the lowest represented class appears roughly 0.4\% times as much as the highest represented category.
We proceed to operate a rebalancing of the dataset by undersampling overrepresented classes.
We decided to pick the 35\textsuperscript{th} percentile of the distribution of class counts as a threshold for considering a category as being \emph{overrepresented}.
We then proceeded to undersample all categories above this threshold by discarding entries containing only instances of overrepresented classes, prioritizing the most common ones for removal.
After this process, we obtained a class distribution as in \Cref{fig:punches_distribution}c.

\subsection{Object Detection with YOLOv10}

As introduced in \cref{sec:intro}, OD operates a recognition of the single instances of objects of known categories within images.
YOLO \cite{redmon2016you} is a one-shot object detector, i.e., it simultaneously predicts object categories and their location within an image.
It conceptually divides an input image into a $S\times S$ grid and outputs a fixed number of \emph{candidate} predictions for each of the elements in the grid.
The candidates are produced even in areas of the model where there may not be any instance of known objects.
Each prediction contains information about the coordinates of the bouding boxes, a \emph{confidence score} encoding the likelihood that the proposed bounding box contains an object, and the \emph{class probabilities}, which indicate the probability of the bounding box being classified into each category.

In the present work, we make use of YOLOv10 \cite{wang2024yolov10}, a modern YOLO architecture which processes input images through a feature extraction \emph{backbone} composed of convolutional and attention layers, leading to three \emph{detection heads}, which are tasked with outputting predictions at a different scale, allowing the model to recognize objects at different scales and sizes.
The main difference introduced by YOLOv10 is the absence of NMS, which was used in previous versions to de-duplicate redundant predictions.
NMS is instead supplanted by a one-to-many and one-to-one prediction matching---called Dual Label Assignment---, which achieves functionally similar results to NMS while being faster to compute.
Other differences include architectural modifications, hyperparameter tuning, and other methodological updates that add incremental performance, both in terms of runtime efficiency and prediction accuracy, with respect to previous YOLO versions.
The reasons behind the adoption of YOLOv10 are twofold: on the one hand, YOLO-like architectures have demonstrated promising performance when it comes to the detection of objects within the artistic domain. For example, Sabatelli et al.\ \cite{sabatelli2021advances} used the popular YOLOv3 version of the algorithm to benchmark musical instrument detection within their newly introduced MINERVA dataset, while an improved version of the algorithm was later used by Wang et al.\ \cite{wang2022object} for detecting paint surface defects. 
On the other hand, the choice of this architecture is also driven by more practical considerations; it is well-known to perform accurately when it comes to large-size images, like the frames considered in this study, and is overall also easy to implement given its off-the-shelf availability and support.

\subsubsection{Adapting YOLOv10 for large-image Object Detection}

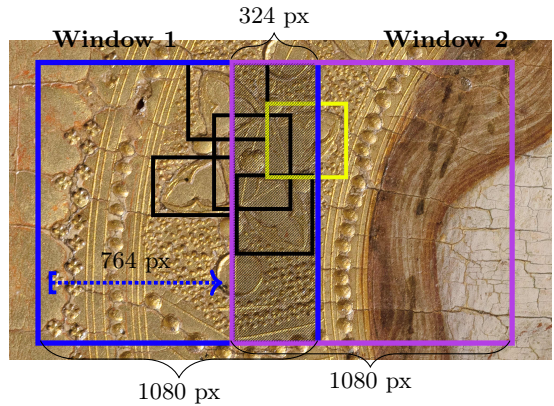
\begin{figure}[t]
    \centering
    \input{img/slidingwindow}
    \caption{
    Illustration of the sliding window approach we adopted for the inference phase of the YOLOv10 model.
    A region of the painting presents several punchmarks, bounded with \textbf{black} boxes.
    We crop a frame of size $1088 \times 1088$ px and apply YOLOv10 on this frame.
    We slide the window by 764 px and obtain new predictions.
    Notice how the punchmark in the \textcolor{darkyellow}{\textbf{yellow}} box, which was partially contained in window 1, is now entirely contained in window 2.
    }
    \label{fig:yolo-nas-inference}
\end{figure}

YOLOv10 supports inference on images of size up to \num{1088} px per side, thus making it unfeasible to apply it to the unprocessed images in our dataset.
As introduced in \Cref{sec:dataset}, we solve this issue for the training procedure by decomposing the large-scale images into \num{70000} frames of size $1088\times 1088$.
With reference to the test image, instead, we resort to applying YOLOv10 using a sliding window approach.
Starting from the top-left corner, we apply YOLOv10 to the first $1088\times 1088$ frame, then slide the window by \num{770} px, and finally apply YOLOv10 again to that window.
The reason for the \num{324} px overlap lies in the fact that we wish to avoid punchmarks being split between two windows.
We illustrate this procedure in \Cref{fig:yolo-nas-inference}.
Since \num{324} is the biggest side in the ground truth bounding boxes in our dataset, we set the overlap to this value.
By introducing an overlap, however, we increase the chance of YOLOv10 outputting multiple predictions referring to the same punch mark instance in two different windows.
For solving this issue, we propose to combine the predictions throughout all windows, then we apply a round of a custom NMS algorithm to get rid of overlapping predictions.

\paragraph{Custom NMS}

YOLO NMS algorithm works by identifying potentially duplicated predictions within the same area using Intersection-over-Union (IoU) and then removing the least confident predictions within this area.
Given two bounding boxes $B_1, B_2$, IoU is defined as $(B_1 \cap B_2)/(B_1 \cup B_2).$

\begin{figure}[t]
    \centering
    \includegraphics[width=0.35\linewidth]{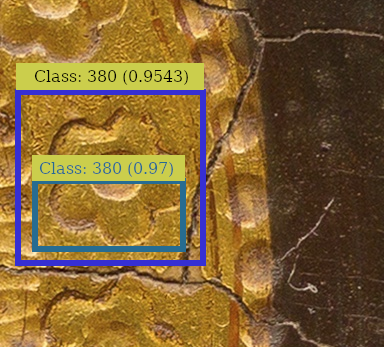}
    \caption{Exemplification of a common consequence of the merging procedure.
    Two high-confidence nested predictions coming from two different frames are both connected to the same punch mark.
    YOLO NMS would keep the smaller one due to its higher confidence.
    }
    \label{fig:nms_motivation}
\end{figure}

\begin{algorithm}[H]
\caption{Custom NMS with IoM}
\label{algo:nms}
\KwIn{List of bounding boxes $b$ of size $N$ and corresponding predicted categories $k$ and confidence scores $c$; IoM threshold $t$; confidence threshold $c^\star$.}
\KwOut{$r$, set of indices to remove.}
$b, k, c \gets \texttt{filter\_confidence}(b, k, c, c^\star)$\; 
$r \gets \{~\}$\;
$M \gets \texttt{pairwise\_IoM}(b)$\; 
\For{$i \gets 1$ \KwTo $N$}{
    \If{$i \notin r$}{
        $v \gets M.\texttt{filter}(\geq t, k[i])$\; 
        $a \gets \texttt{compute\_areas}(b[v])$\; 
        $v.\texttt{sort}(a)$\; 
        $v.\texttt{pop}(1)$\; 
        $r.\texttt{add}(v)$\; 
    }
}
\Return $r$\;
\end{algorithm}
In our case, after merging the predictions for multiple images, we are left with many cases of nested predictions, as exemplified in \Cref{fig:nms_motivation}.
Applying YOLO NMS may cause a large number of small, high-confidence predictions to coalesce over larger ones.
This would lead to a lower localization accuracy of the model due to a low IoU between ground truth boxes and predictions.
We modify the NMS algorithm first by adding a phase in which we remove predictions with confidence lower than a threshold $c^\star$, then by replacing IoU with the Intersection-over-Minimum (IoM) \cite{vogel2023fully}:
$$
    \text{IoM}(B_1, B_2) = \frac{B_1 \cap B_2}{\min\{B_1,B_2\}}.
$$

We then calculate the pairwise IoM between the boxes.
We coalesce same-class predictions with IoM above a certain threshold by removing all predictions but the one with the largest area. A pseudocode version of the algorithm is presented in Algorithm \ref{algo:nms}.




\subsection{Evaluation metrics}

In order to assess the task-level performance of our model, we make use of the following popular metrics for OD tasks:

\textbf{Precision} is computed as the ratio between True Positives (TPs) and all of the model predictions.
In OD, a TP is defined as a prediction whose bounding box intersects a corresponding ground truth bounding box belonging to the same category.
We consider the two boxes to match when IoU is larger than a threshold $\tau\in[0,1]$.
We report Precision with $\tau=0.5$ (\texttt{P@.5}).
Precision highlights the model's capability of correctly classifying punches, but it ignores False Negatives---i.e., ground truth bounding boxes with no matching prediction.



\textbf{Recall} is calculated as the ratio between TP and all false negatives---i.e., ground truth objects with no matching prediction.
We report Recall at the IoU threshold $\tau$ of 0.5 (\texttt{R@.5}).
It highlights the model's capability of exhaustively identifying punches within an image, but it ignores incorrect classifications.
Precision and Recall can be combined via harmonic mean to provide the \textbf{F1-Score}.
This metric considers both the capability of the model to output correct predictions and reduce false negatives.
We report F1-Score at the IoU threshold $\tau$ of 0.5 \texttt{F1@.5}. For validation purposes only, we consider the \textbf{mean Average Precision} (mAP) metric, that summarizes the ability of the model output correct predictions at different confidence levels and IoU thresholds.
We do not make use of mAP in the test-set evaluation since the NMS procedure already removes predictions below a confidence threshold, thus rendering useless the necessity for calculating AP at different confidence levels.
In addition, metrics such as Precision, Recall, and F1-Score are easier to communicate to model stakeholders (e.g., art historians) who may not be expert in machine learning or statistics.



\subsection{Experimental settings}

\begin{table}[t]
\caption{Model size and latency of the YOLOv10 variants used in our work, as reported by Wang et al.\ \cite{wang2024yolov10}.
    We define relative latency as the latency reported in their work over the latency of their fastest model (YOLOv10n).
    }
    \centering
    \begin{tabular}{l@{\hskip 0.5cm}c@{\hskip 0.5cm}c}
        \toprule
        Model name & Number of parameters & Relative latency \\
        \midrule
        YOLOv10n & $2.3\times 10^6$ & 1.00 \\
        YOLOv10s & $7.2\times 10^6$ & 1.35 \\
        YOLOv10l & $24.4\times 10^6$ & 3.96 \\
        \bottomrule
    \end{tabular}

    \label{tab:yolo_versions}
\end{table}

For training YOLOv10 on our dataset of punches, we make use of the following three variants: YOLOv10n, YOLOv10s, and YOLOv10l.
The only difference between these three models is the number of parameters, which we present in \Cref{tab:yolo_versions}.
As is common practice when dealing with datasets that are far in terms of size from the ones that are typically used as benchmarks by the computer-vision community, we rely on a transfer-learning approach. Following the guidelines presented in \cite{sabatelli2018deep}, we started from a model pre-trained on the Common Objects in COntext (COCO) dataset \cite{lin2014microsoft}, which we fine-tuned for 100 epochs using regular Stochastic Gradient Descent with a batch size of 16, an initial learning rate of 0.01, momentum of 0.9, and weight decay of 0.0005, using the classical YOLO loss \cite{redmon2016you}. Throughout each epoch, we performed a model assessment on the validation set by computing \texttt{mAP@.5:.9}. We performed early stopping \cite{prechelt2002early} by selecting, at the end of the training, the parameters with the best validation performance.
We tuned the hyperparameters confidence threshold $c^\star \in [0.5, 0.8]$ and IoM threshold $t \in [0.5, 0.95]$ by running the sliding window algorithm with custom NMS on the non-testing images and selecting the combination which yielded the highest \texttt{mAP@.5:.9}.
We determined the best combination to be $c^\star=0.75$ and $t=0.7$ for YOLOv10n, $c^\star=0.8$ and $t=0.6$ for YOLOv10s, and $c^\star=0.8$ and $t=0.5$ for YOLOv10l. We performed all experiments with Python version 3.9.20 and the PyTorch library \cite{paszke2017automatic} version 2.0.1 with CUDA 11.7 on an NVIDIA A100 GPU with 40 GB of VRAM.

\section{Results}

The results attained by the models on the validation split are presented in \Cref{tab:results_val}, while \Cref{tab:results1} showcases a per-class overview of the results on the held-out picture.
The model YOLOv10n is the one with the best results in terms of F1-Score, which hints at the possibility that the task may not need an extreme level of overparameterization to be tackled.
While the Precision of all three models is around 94\%, YOLOv10n has a much better Recall (89.81\%, compared to 86.47\% of YOLOv10s and 78.49\% of YOLOv10l), showcasing how larger models struggle more with false negatives.
It needs to be noticed that the high Precision translates also to having very few instances of predictions whose category is not present in the specific picture (2 predictions out of 1690 for YOLOv10n, 1 out of 1625 for YOLOv10s, and none for YOLOv10l).
This is important from an art historian perspective, since he/she prefers a lower Recall to having the predictions \emph{polluted} with false positives, which may point to different punches being used in the painting, and hence hint at different authors than the expectation.

Additionally, we can see how the custom NMS procedure boosts the Precision and F1-Score of YOLOv10n and YOLOv10s at the expense of Recall (albeit at a smaller magnitude than Precision).
This behavior does not instead occur in YOLOv10l, which apparently struggles more with outputting high-confidence accurate predictions.
The Precision boost observed is expected since the criterion for selecting the best configuration of parameters for NMS is based on mAP: a high confidence threshold will increase true positives, but introduce more false negatives, hence the behavior observed in the table.
The drop in Recall is particularly noticeable for punch \#333, whose Recall in YOLOv10s reached a low of 48.78\% and even as low as 15.45\% in YOLOv10l, meaning that the model missed the majority of its instances in the picture.

\begin{table}[t]
    \centering
    \caption{Results in terms of Precision, Recall, F1-Score, and mAP achieved by our three models on the validation dataset at the early stopping epoch.}
    \begin{tblr}{
        colspec = {l*{4}{r}},
        hline{1,5} = {-}{0.08em},
        hline{2} = {-}{0.02em}
    }
        Model & \texttt{P@.5} & \texttt{R@.5} & \texttt{F1@.5} & mAP \\
        YOLOv10n & 0.813 & 0.707 & 0.770 & 0.590 \\
        YOLOv10s & 0.827 & 0.694 & 0.755 & 0.584 \\
        YOLOv10l & 0.811 & 0.652 & 0.723 & 0.557 \\
    \end{tblr}
    
    \label{tab:results_val}
\end{table}

\begin{table}[t]
\caption{Per-punch category results in terms of Precision, Recall, and F1-Score achieved by our three models on the test datraset before and after the application of our custom NMS routine.
    The last three columns indicate the percent variation in the metrics ($\mathrm{\Delta}_\%$) ascribable to NMS.
    The category ``others'' refers to punch classes present in the predictions but not in the ground truth labels.}
    \centering
    \resizebox{\textwidth}{!}{%
\begin{tblr}{
  colspec = {l*{10}{r}},
  hline{1,20} = {-}{0.125em},
  hline{9,15} = {-}{0.1em},
  hline{3,8,14,19} = {-}{0.02em},
  vline{3} = {3-Z}{0.02em},
  vline{7} = {3-Z}{0.02em},
  vline{11} = {3-Z}{0.02em},
  cell{1}{3} = {c=4}{c},
  cell{1}{7} = {c=4}{c},
  cell{1}{11} = {c=3}{c},
  cell{3}{1} = {r=6}{c},
  cell{9}{1} = {r=6}{c},
  cell{15}{1}= {r=5}{c},
  font=\footnotesize
  }
& &   Before NMS & & &   & After NMS & &  & & $\mathrm{\Delta}_\%$ & & \\
& Cat. & n & \texttt{P@.5} & \texttt{R@.5} & \texttt{F1@.5} & n &\texttt{P@.5} & \texttt{R@.5} &  \texttt{F1@.5} & \texttt{P@.5} & \texttt{R@.5} & \texttt{F1@.5} \\
\rotatebox{90}{YOLOv10n} & 47 & 532 & 0.7274 & 0.9748 & 0.8332 & 210 & 0.9381 & 0.9517 & 0.9448 & 22.46\% & -2.43\% & 11.81\% \\
&138 & 614 & 0.6889 & 0.9883 & 0.8119 & 210 & 0.9095 & 0.9745 & 0.9409 & 24.26\% & -0.14\% & 13.71\% \\
&333 & 138 & 0.7826 & 0.6750 & 0.7248 & 74 & 0.9595 & 0.5772 & 0.7208 & 18.44\% & -16.94\% & -0.55\% \\
&388 & 404 & 0.9183 & 0.9027 & 0.9104 & 197 & 0.9797 & 0.8283 & 0.8977 & 6.27\% & -8.98\% & -1.41\% \\
&others & 2 & 0.0000 & -- & -- & 2 & 0.0000 & -- & -- & -- & -- & -- \\
&ALL & 1690 & 0.7627 & \textbf{0.9234} & 0.8354 & 693 & 0.9408 & \textbf{0.8590} & \textbf{0.8981} & 18.93\% & \textbf{-7.50\%} & \textbf{6.98\%} \\

\rotatebox{90}{YOLOv10s} & 47 & 470 & 0.7191 & 0.9160 & 0.8057 & 185 & 0.9405 & 0.8406 & 0.8878 & 23.54\% & -8.97\% & 9.25\% \\
&138 & 615 & 0.6992 & 0.9931 & 0.8206 & 205 & 0.9415 & 0.9847 & 0.9626 & 25.74\% & -0.85\% & 14.75\% \\
&333 & 139 & 0.7770 & 0.6585 & 0.7129 & 64 & 0.9375 & 0.4878 & 0.6417 & 17.12\% & -34.99\% & -11.10\% \\
&388 & 400 & 0.8575 & 0.8728 & 0.8651 & 183 & 0.9672 & 0.7597 & 0.8510 & 11.34\% & -14.89\% & -1.66\% \\
&others & 1 & 0.0000 & -- & -- & 1 & 0.0000 & -- & -- & -- & -- & -- \\
&ALL & 1625 & 0.7502 & 0.8970 & 0.8170 & 638 & \textbf{0.9467} & 0.7958 & 0.8647 & \textbf{20.76\%} & -12.72\% & 5.52\% \\

\rotatebox{90}{YOLOv10l} & 47 & 192 & 0.7962 & 0.8032 & 0.7997 & 150 & 0.9400 & 0.6812 & 0.7899 & 15.30\% & -17.91\% & -1.24\% \\
&138 & 165 & 0.7022 & 0.7957 & 0.7460 & 198 & 0.9444 & 0.9541 & 0.9492 & 25.65\% & 16.60\% & 21.41\% \\
&333 & 2 & 0.7817 & 0.9790 & 0.8693 & 28 & 0.6786 & 0.1545 & 0.2517 & -15.19\% & -533.66\% & -245.37\% \\
&388 & 29 & 0.5897 & 0.1811 & 0.2771 & 167 & 0.9820 & 0.7039 & 0.8200 & 39.95\% & 74.27\% & 66.21\% \\
&ALL & 1625 & \textbf{0.9485} & 0.8194 & \textbf{0.8792} & 543 & 0.9411 & 0.6733 & 0.7849 & -0.79\% & -21.70\% & -12.01\% \\
\end{tblr}
}
    
    \label{tab:results1}
\end{table}

\section{Discussion and Conclusions}

In the current work, we presented a YOLOv10-based pipeline for operating predictions on punchmarks in images of panel paintings in the Florence area in the late Middle Ages.
We first obtained, following a thorough photographic setting, a dataset of a few large-scale images of 8 paintings, which we proceeded to manually label, identifying \num{3745} occurrences of punchmarks across 27 categories.
We then extracted frames from these pictures by means of random subwindows of size $1088\times1088$.
Due to the very large class imbalance, we rebalanced by subsampling majority classes.
Finally, we split the frames into training and validation, carefully avoiding leakage.
We then proceeded to train three variants of YOLOv10.
In order to combine the predictions operated on small frames onto the bigger pictures in our dataset, we resorted to a sliding window approach, overlapping each window to avoid splitting punches between different windows.
When combining the predictions, we needed to take into consideration possible multiple duplicate high-confidence predictions coming from different windows.
We tackled this issue with a custom non-maximal suppression strategy making use of the Intersection-over-Minimum metric.
We showed, on a large-scale image held out in our dataset, how YOLOv10 is capable of producing highly precise predictions.
In addition, we showed how our custom NMS strategy is capable of increasing the accuracy of the predictions output by two out of three of our YOLOv10 models in terms of Precision and F1 score, limiting the decrease in Recall.

Despite these results, our study still has various limitations.
First of all, we must notice our dataset is still composed of a low number of paintings, which hinders especially the evaluation phase.
A test dataset including more picture and covering more punch classes---especially those underrepresented in the training dataset---would provide with the possibility of evaluating the models outside of the single 4 punches categories from \Cref{tab:results1}.
However, given the expensive labor of the pictures shooting and the manual labeling procedures, this is an arduous task to carry out.
An additional goal could be to use data (both for training and evaluation) coming from badly preserved paintings: this would allow test more difficult cases, and would also support the evaluation of the model on out-of-distribution data, similarly to what done previously by Zullich et al.\ \cite{zullich2023artificial}.
Finally, we made use of some YOLOv10 variants: despite their good trade-off between detection speed and accuracy, two-stage models could provide more accurate results.
Moreover, our training could be performed with additional configurations of hyperparameters and optimizers which may yield better results.

For what concerns future improvements over the current pipeline, we imagine our work to be of interest to users who may want to apply our model while working on the field---in this sense, it would be beneficial to train the model on pictures obtained in a less professional setting and at different scales.
In addition, the model predictions may be crossed with the existing knowledge base provided by Skaug in his catalogs \cite{skaug1994punch} to automatically notify the users about predictions that violate this knowledge.

All in all, we believe our work to be an initial step in the direction of providing art historians with an automatic tool to help with author attribution in a quantitative and scientific way.

\section{Acknowledgements}

We thank the Center for Information Technology of the University of Groningen for their support and for providing access to the Hábrók high performance computing cluster.


%
%
%
\bibliographystyle{splncs04}
\bibliography{biblio}

\end{document}

%% file: plots/data_barchart.tex
\begin{tikzpicture}
    \begin{axis}[
        width=4cm, 
        height=8cm,
        xlabel={Count},
        ylabel={Punch id},
        ytick={1,2,3,4,5,6,7,8,9,10,11,12,13,14,15,16,17,18,19,20,21,22,23,24,25,26,27},
        yticklabels={388,	47,	392,	360,	333,	138,	649,	537,	606,	631,	380,	558,	690,	418,	644,	377,	69,	403,	183,	2,	652,	599,	680,	196,	405,	376,	233,
},
        xbar,
        bar width=0.15cm,
        nodes near coords,
        nodes near coords style={
            anchor=west, 
            black, 
            xshift=2pt, 
        },
        point meta=explicit symbolic,
        enlargelimits=0.1,
        axis lines=left,
        axis line style=-,
        y dir=reverse,
        tick align=outside,
        clip=false,
        ymax=27.5,
        font=\scriptsize 
    ]
    \addplot coordinates {
        ( 687 , 1 ) [687]
        ( 434 , 2 ) [434]
        ( 432 , 3 ) [432]
        ( 363 , 4 ) [363]
        ( 344 , 5 ) [344]
        ( 317 , 6 ) [317]
        ( 175 , 7 ) [175]
        ( 144 , 8 ) [144]
        ( 127 , 9 ) [127]
        ( 123 , 10 ) [123]
        ( 102 , 11 ) [102]
        ( 95 , 12 ) [95]
        ( 85 , 13 ) [85]
        ( 51 , 14 ) [51]
        ( 36 , 15 ) [36]
        ( 35 , 16 ) [35]
        ( 31 , 17 ) [31]
        ( 29 , 18 ) [29]
        ( 28 , 19 ) [28]
        ( 27 , 20 ) [27]
        ( 21 , 21 ) [21]
        ( 17 , 22 ) [17]
        ( 12 , 23 ) [12]
        ( 10 , 24 ) [10]
        ( 10 , 25 ) [10]
        ( 6 , 26 ) [6]
        ( 4 , 27 ) [4]

    };
    \end{axis}
    \node at (-0.2,-0.75) {(a)};
    
\end{tikzpicture}

%% file: plots/data_barchart_before_rebalancing.tex
\begin{tikzpicture}
    \begin{axis}[
        width=4cm, 
        height=8cm,
        xlabel={Count},
        ytick={1,2,3,4,5,6,7,8,9,10,11,12,13,14,15,16,17,18,19,20,21,22,23,24,25,26,27},
        yticklabels={360, 	388, 	333, 	690, 	537, 	47, 	392, 	631, 	380, 	649, 	138, 	606, 	558, 	69, 	644, 	377, 	652, 	403, 	2, 	680, 	405, 	183, 	196, 	599, 	418, 	376, 	233
},
        xbar,
        bar width=0.15cm,
        nodes near coords,
        nodes near coords style={
            anchor=west, 
            black, 
            xshift=2pt, 
        },
        point meta=explicit symbolic,
        enlargelimits=0.1,
        axis lines=left,
        axis line style=-,
        y dir=reverse,
        tick align=outside,
        clip=false,
        ymax=27.5,
        font=\scriptsize 
    ]
    \addplot coordinates {
         (39293,1) [\num{39293}]
        (27374,2) [\num{27374}]
        (21676,3) [\num{21676}]
        (19165,4) [\num{19165}]
        (16037,5) [\num{16037}]
        (15823,6) [\num{15823}]
        (15114,7) [\num{15114}]
        (11908,8) [\num{11908}]
        (8911,9) [\num{8911}]
        (8841,10) [\num{8841}]
        (7125,11) [\num{7125}]
        (4429,12) [\num{4429}]
        (3891,13) [\num{3891}]
        (3423,14) [\num{3423}]
        (3412,15) [\num{3412}]
        (3237,16) [\num{3237}]
        (3099,17) [\num{3099}]
        (2675,18) [\num{2675}]
        (1543,19) [\num{1543}]
        (1476,20) [\num{1476}]
        (1348,21) [\num{1348}]
        (1154,22) [\num{1154}]
        (1022,23) [\num{1022}]
        (719,24) [\num{719}]
        (399,25) [\num{399}]
        (288,26) [\num{288}]
        (178,27) [\num{178}]

    };
    \end{axis}
    \node at (-0.2,-0.75) {(b)};
\end{tikzpicture}

%% file: plots/data_barchart_after_rebalancing.tex
\begin{tikzpicture}
    \begin{axis}[
        width=4cm, 
        height=8cm,
        xlabel={Count},
        scaled x ticks=base 10:-3,
        ytick={1,2,3,4,5,6,7,8,9,10,11,12,13,14,15,16,17,18,19,20,21,22,23,24,25,26,27},
        yticklabels={47, 	690, 	631, 	644, 	69, 	360, 	333, 	392, 	380, 	138, 	377, 	388, 	606, 	405, 	537, 	649, 	652, 	403, 	2, 	680, 	405, 	183, 	196, 	599, 	418, 	376, 	233
},
        xbar,
        bar width=0.15cm,
        nodes near coords,
        nodes near coords style={
            anchor=west, 
            black, 
            xshift=2pt, 
        },
        point meta=explicit symbolic,
        enlargelimits=0.1,
        axis lines=left,
        axis line style=-,
        y dir=reverse,
        tick align=outside,
        clip=false,
        ymax=27.5,
        font=\scriptsize 
    ]
    \addplot coordinates {
         (8665,1) [\num{8665}]
        (7457,2) [\num{7457}]
        (6708,3) [\num{6708}]
        (3254,4) [\num{3254}]
        (3216,5) [\num{3216}]
        (2829,6) [\num{2829}]
        (2717,7) [\num{2717}]
        (2717,8) [\num{2717}]
        (2717,9) [\num{2717}]
        (2717,10) [\num{2717}]
        (2717,11) [\num{2717}]
        (2716,12) [\num{2716}]
        (2716,13) [\num{2716}]
        (2716,14) [\num{2716}]
        (2715,15) [\num{2715}]
        (2715,16) [\num{2715}]
        (2712,17) [\num{2712}]
        (2675,18) [\num{2675}]
        (1543,19) [\num{1543}]
        (1476,20) [\num{1476}]
        (1348,21) [\num{1348}]
        (1154,22) [\num{1154}]
        (1022,23) [\num{1022}]
        (719,24) [\num{719}]
        (399,25) [\num{399}]
        (288,26) [\num{288}]
        (178,27) [\num{178}]

    };
    \end{axis}
    \node at (-0.2,-0.75) {(c)};
\end{tikzpicture}

%% file: img/slidingwindow.tex
\begin{tikzpicture}[scale=1.1]
    \node at (0,0) {\includegraphics[scale=1.1]{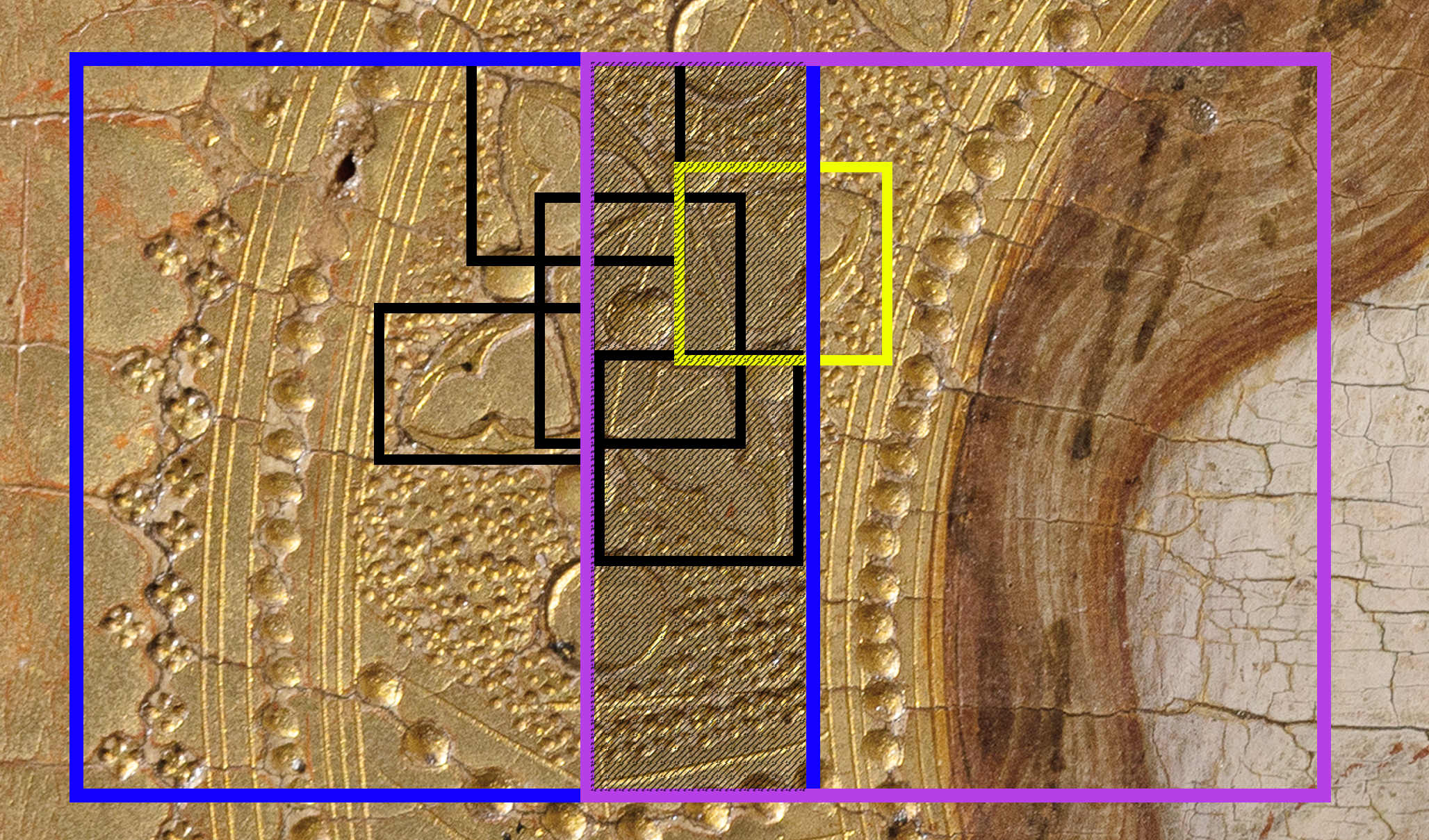}};
    \node at (-2,1.95) {\textbf{Window 1}};
    \node at (2,1.95) {\textbf{Window 2}};
    \draw[Bracket->, line width=0.5mm, color=blue, densely dotted] (-2.8,-1) -- (-0.7,-1) node[midway, above, black] {764 px};
    \draw [decorate, decoration={brace, amplitude=14pt, mirror}] (-2.9,-1.7) -- (0.45,-1.7)
        node[midway, yshift=-20pt] {1080 px};
    \draw [decorate, decoration={brace, amplitude=10pt, mirror}] (-0.6,-1.7) -- (2.775,-1.7)
        node[midway, yshift=-16pt] {1080 px};
    \draw [decorate, decoration={brace, amplitude=10pt}] (-0.6,1.7) -- (0.45,1.7)
        node[midway, yshift=16pt] {324 px};
\end{tikzpicture}

%% file: main.bbl
\begin{thebibliography}{10}
\providecommand{\url}[1]{\texttt{#1}}
\providecommand{\urlprefix}{URL }
\providecommand{\doi}[1]{https://doi.org/#1}

\bibitem{ansari2021survey}
Ansari, M., Lodi, K.: A survey of recent trends in two-stage object detection methods. In: Renewable Power for Sustainable Growth: Proceedings of International Conference on Renewal Power (ICRP 2020). pp. 669--677. Springer (2021)

\bibitem{bar2015classification}
Bar, Y., Levy, N., Wolf, L.: Classification of artistic styles using binarized features derived from a deep neural network. In: Computer Vision-ECCV 2014 Workshops: Zurich, Switzerland, September 6-7 and 12, 2014, Proceedings, Part I 13. pp. 71--84. Springer (2015)

\bibitem{bengamraComprehensiveSurveyObject2024}
Bengamra, S., Mzoughi, O., Bigand, A., Zagrouba, E.: A comprehensive survey on object detection in {{Visual Art}}: Taxonomy and challenge. Multimedia Tools and Applications  \textbf{83}(5),  14637--14670 (Feb 2024). \doi{10.1007/s11042-023-15968-9}

\bibitem{carion2020end}
Carion, N., Massa, F., Synnaeve, G., Usunier, N., Kirillov, A., Zagoruyko, S.: End-to-end object detection with transformers. In: European conference on computer vision. pp. 213--229. Springer (2020)

\bibitem{carranza2020performance}
Carranza-Garc{\'\i}a, M., Torres-Mateo, J., Lara-Ben{\'\i}tez, P., Garc{\'\i}a-Guti{\'e}rrez, J.: On the performance of one-stage and two-stage object detectors in autonomous vehicles using camera data. Remote Sensing  \textbf{13}(1), ~89 (2020)

\bibitem{cetinic2018fine}
Cetinic, E., Lipic, T., Grgic, S.: Fine-tuning convolutional neural networks for fine art classification. Expert Systems with Applications  \textbf{114},  107--118 (2018)

\bibitem{corridoni1996visual}
Corridoni, J.M., Del~Bimbo, A., De~Magistris, S., Vicario, E.: A visual language for color-based painting retrieval. In: Proceedings 1996 IEEE Symposium on Visual Languages. pp. 68--75. IEEE (1996)

\bibitem{davidDeepPainterPainterClassification2016}
David, O.E., Netanyahu, N.S.: {{DeepPainter}}: {{Painter Classification Using Deep Convolutional Autoencoders}}. In: Villa, A.E., Masulli, P., Pons~Rivero, A.J. (eds.) Artificial {{Neural Networks}} and {{Machine Learning}} -- {{ICANN}} 2016, vol.~9887, pp. 20--28. Springer International Publishing, Cham (2016). \doi{10.1007/978-3-319-44781-0_3}

\bibitem{frinta1972punched}
Frinta, M.S.: On the punched decoration in {M}edieval panel painting and manuscript illumination. Studies in Conservation  \textbf{17}(sup1),  115--121 (1972)

\bibitem{girshick2014rich}
Girshick, R., Donahue, J., Darrell, T., Malik, J.: Rich feature hierarchies for accurate object detection and semantic segmentation. In: Proceedings of the IEEE conference on computer vision and pattern recognition. pp. 580--587 (2014)

\bibitem{gonthier2022multiple}
Gonthier, N., Ladjal, S., Gousseau, Y.: Multiple instance learning on deep features for weakly supervised object detection with extreme domain shifts. Computer Vision and Image Understanding  \textbf{214},  103299 (2022)

\bibitem{hachimura1996retrieval}
Hachimura, K.: Retrieval of paintings using principal color information. In: Proceedings of 13th International Conference on Pattern Recognition. vol.~3, pp. 130--134. IEEE (1996)

\bibitem{johnsonImageProcessingArtist2008}
Johnson, C.R., Hendriks, E., Berezhnoy, I.J., Brevdo, E., Hughes, S.M., Daubechies, I., Li, J., Postma, E., Wang, J.Z.: Image processing for artist identification. IEEE Signal Processing Magazine  \textbf{25}(4),  37--48 (Jul 2008). \doi{10.1109/MSP.2008.923513}

\bibitem{kronerAuthenticationFreeHand1998}
Kr\"{o}ner, S., Lattner, A.: Authentication of free hand drawings by pattern recognition methods. In: Proceedings. {{Fourteenth International Conference}} on {{Pattern Recognition}} ({{Cat}}. {{No}}.{{98EX170}}). vol.~1, pp. 462--464. {IEEE Comput. Soc}, {Brisbane, Qld., Australia} (1998). \doi{10.1109/ICPR.1998.711180}

\bibitem{lettner2004texture}
Lettner, M., Kammerer, P., Sablatnig, R.: Texture analysis of painted strokes. na (2004)

\bibitem{lin2014microsoft}
Lin, T.Y., Maire, M., Belongie, S., Hays, J., Perona, P., Ramanan, D., Doll{\'a}r, P., Zitnick, C.L.: Microsoft coco: Common objects in context. In: Computer Vision--ECCV 2014: 13th European Conference, Zurich, Switzerland, September 6-12, 2014, Proceedings, Part V 13. pp. 740--755. Springer (2014)

\bibitem{maree2004generic}
Mar{\'e}e, R., Geurts, P., Piater, J., Wehenkel, L.: A generic approach for image classification based on decision tree ensembles and local sub-windows. In: 6th Asian Conference on Computer Vision. Asian Federation of Computer Vision Societies (AFCV) (2004)

\bibitem{melzerStrokeDetectionBrush1998}
Melzer, T., Kammerer, P., Zolda, E.: {Stroke Detection of Brush Strokes in Portrait Miniatures Using a Semi-Parametric and a Model Based Approach}. In: Proceedings. {{Fourteenth International Conference}} on {{Pattern Recognition}} ({{Cat}}. {{No}}.{{98EX170}}). vol.~1, pp. 474--476 vol.1 (1998). \doi{10.1109/ICPR.1998.711184}

\bibitem{milanifedericoDatasetConvolutionalModel2021}
Milani, F., Fraternali, P.: A {{Dataset}} and a {{Convolutional Model}} for {{Iconography Classification}} in {{Paintings}}. Journal on Computing and Cultural Heritage (JOCCH)  (Jul 2021). \doi{10.1145/3458885}

\bibitem{nanni2014ensemble}
Nanni, L., Lumini, A., Brahnam, S.: Ensemble of different local descriptors, codebook generation methods and subwindow configurations for building a reliable computer vision system. Journal of King Saud University-Science  \textbf{26}(2),  89--100 (2014)

\bibitem{paszke2017automatic}
Paszke, A., Gross, S., Chintala, S., Chanan, G., Yang, E., DeVito, Z., Lin, Z., Desmaison, A., Antiga, L., Lerer, A.: Automatic differentiation in pytorch  (2017)

\bibitem{prechelt2002early}
Prechelt, L.: Early stopping-but when? In: Neural Networks: Tricks of the trade, pp. 55--69. Springer (2002)

\bibitem{redmon2016you}
Redmon, J.: You only look once: Unified, real-time object detection. In: Proceedings of the IEEE conference on computer vision and pattern recognition (2016)

\bibitem{ren2016faster}
Ren, S., He, K., Girshick, R., Sun, J.: Faster r-cnn: Towards real-time object detection with region proposal networks. IEEE transactions on pattern analysis and machine intelligence  \textbf{39}(6),  1137--1149 (2016)

\bibitem{ren2023strong}
Ren, T., Yang, J., Liu, S., Zeng, A., Li, F., Zhang, H., Li, H., Zeng, Z., Zhang, L.: A strong and reproducible object detector with only public datasets. arXiv preprint arXiv:2304.13027  (2023)

\bibitem{ross2017focal}
Ross, T.Y., Doll{\'a}r, G.: Focal loss for dense object detection. In: proceedings of the IEEE conference on computer vision and pattern recognition. pp. 2980--2988 (2017)

\bibitem{sabatelli2021advances}
Sabatelli, M., Banar, N., Cocriamont, M., Coudyzer, E., Lasaracina, K., Daelemans, W., Geurts, P., Kestemont, M.: Advances in digital music iconography: Benchmarking the detection of musical instruments in unrestricted, non-photorealistic images from the artistic domain. Digital Humanities Quarterly  \textbf{15}(1) (2021)

\bibitem{sabatelli2018deep}
Sabatelli, M., Kestemont, M., Daelemans, W., Geurts, P.: Deep transfer learning for art classification problems. In: Proceedings Of The European conference on computer vision (ECCV) workshops. pp.~0--0 (2018)

\bibitem{santosArtificialNeuralNetworks2021}
Santos, I., Castro, L., {Rodriguez-Fernandez}, N., {Torrente-Pati{\~n}o}, {\'A}., Carballal, A.: Artificial {{Neural Networks}} and {{Deep Learning}} in the {{Visual Arts}}: A review. Neural Computing and Applications  \textbf{33}(1),  121--157 (Jan 2021). \doi{10.1007/s00521-020-05565-4}

\bibitem{seguin2016visual}
Seguin, B., Striolo, C., diLenardo, I., Kaplan, F.: Visual link retrieval in a database of paintings. In: Computer Vision--ECCV 2016 Workshops: Amsterdam, The Netherlands, October 8-10 and 15-16, 2016, Proceedings, Part I 14. pp. 753--767. Springer (2016)

\bibitem{shamirComputerAnalysisReveals2012}
Shamir, L.: Computer {{Analysis Reveals Similarities}} between the {{Artistic Styles}} of {{Van Gogh}} and {{Pollock}}. Leonardo  (2012). \doi{10.1162/LEON_a_00281}

\bibitem{skaug1983punch}
Skaug, E.S.: Punch marks---{W}hat are they worth? {P}roblems of {T}uscan workshop interrelationships in the mid fourteenth-century: the {O}vile master and {G}iovanni da {M}ilano. In: La pittura nel XIV e XV secolo. Il contributo dell'analisi tecnica alla storia dell'arte, vol.~3, pp. 253--282 (1983)

\bibitem{skaug1994punch}
Skaug, E.S.: Punch marks from Giotto to Fra Angelico: attribution, chronology, and workshop relationships in Tuscan panel painting c. 1330-1430. Volumes 1-2. IIC, Nordic Group, the Norwegian section (1994)

\bibitem{tyler2012search}
Tyler, C.W., Smith, W.A., Stork, D.G.: In search of leonardo: computer-based facial image analysis of renaissance artworks for identifying leonardo as subject. In: Human Vision and Electronic Imaging XVII. vol.~8291, pp. 407--413. SPIE (2012)

\bibitem{vogel2023fully}
Vogel, F.W., Alipek, S., Eppler, J.B., Triesch, J., Bissen, D., Acker-Palmer, A., Rumpel, S., Kaschube, M.: Fully automated detection of dendritic spines in 3d live cell imaging data using deep convolutional neural networks. bioRxiv pp. 2023--01 (2023)

\bibitem{wang2024yolov10}
Wang, A., Chen, H., Liu, L., Chen, K., Lin, Z., Han, J., Ding, G.: Yolov10: Real-time end-to-end object detection. arXiv preprint arXiv:2405.14458  (2024)

\bibitem{wang2022object}
Wang, J., Su, S., Wang, W., Chu, C., Jiang, L., Ji, Y.: An object detection model for paint surface detection based on improved yolov3. Machines  \textbf{10}(4), ~261 (2022)

\bibitem{Yao2018ActiveTT}
Yao, Y., Sfarra, S., Lag{\"u}ela, S., Lag{\"u}ela, S., Ibarra-Castanedo, C., Wu, J.Y., Maldague, X.P.V., Ambrosini, D.: Active thermography testing and data analysis for the state of conservation of panel paintings. International Journal of Thermal Sciences  \textbf{126},  143--151 (2018)

\bibitem{zullich2023artificial}
Zullich, M., Macovaz, V., Pinna, G., Pellegrino, F.A.: An artificial intelligence system for automatic recognition of punches in fourteenth-century panel painting. IEEE Access  \textbf{11},  5864--5883 (2023)

\end{thebibliography}
